\documentclass[11pt,a4paper]{erlarticle}
\usepackage{amsmath,amssymb,mathtools}
\usepackage{booktabs}
\usepackage{graphicx}
\usepackage{subcaption}
\usepackage{float}
\usepackage{placeins}
\usepackage{algorithm}
\usepackage{algpseudocode}
\usepackage[table]{xcolor}
\usepackage[nameinlink,capitalise]{cleveref}
\crefname{figure}{Fig.}{Figs.}
\Crefname{figure}{Fig.}{Figs.}
\makeatletter
\providecommand{\bstctlcite}[1]{%
  \@bsphack
  \if@filesw\immediate\write\@auxout{\string\citation{#1}}\fi
  \@esphack}
\makeatother
\newcommand{\method}{\textsc{WAM-OPD}}

\definecolor{teachergray}{HTML}{65717D}
\definecolor{offlinegray}{HTML}{F1F3F5}
\definecolor{studentgold}{HTML}{FFF1CC}
\definecolor{ourblue}{HTML}{EAF3FB}
\definecolor{ourbluehead}{HTML}{D6E8F6}
\definecolor{gainblue}{HTML}{174A7E}

\title{WAM-OPD: Joint Video--Action Supervision\\
for World Action Model Post-Training\\
with On-Policy Distillation}
\runningtitle{WAM-OPD}
\author[1]{Liuhaichen~Yang}
\author[1,\textdagger]{Zhengyang~Zhong}
\author[1,\textdagger]{Hanshang~Zhu}
\author[1]{Ningwei~Bai}
\author[1]{Qichen~Yin}
\author[2]{Zhi~Han}
\author[1]{Jiarui~Qin}
\author[1]{Zhuang~Jiang}
\author[1]{Chenchao~Sheng}
\author[3]{Hanbo~Ma}
\author[4]{Junkai~Liu}
\author[4]{Junkai~Sun}
\author[4]{Dongcheng~Lyu}
\author[5]{Bo~Liu}
\author[6]{Yi~Dong}
\author[1,*]{Zezhi~Tang}

\affil[1]{\mbox{University College London}\quad
\mbox{\textsuperscript{2}Shanghai Jiao Tong University}\quad
\mbox{\textsuperscript{3}The University of Manchester}\quad
\mbox{\textsuperscript{4}Chinese Academy of Sciences, Nanjing Institute of Software Technology}\quad
\mbox{\textsuperscript{5}Shenzhen University of Advanced Technology}\quad
\mbox{\textsuperscript{6}University of Liverpool}}
\titlenote{\textsuperscript{\textdagger}Equal contribution.\quad
\textsuperscript{*}Corresponding author: Zezhi Tang. E-mail: \texttt{zezhi.tang@ucl.ac.uk}.}
\date{}
\begin{document}
\bstctlcite{IEEEauthorcontrol}
\maketitle
\begin{abstract}
World Action Models (WAMs) generate both future video and robot actions, offering two connected outputs for post-training supervision. How can a pretrained WAM learn from a stronger Teacher on the histories it encounters during execution? We present WAM-OPD, which collects Student rollout histories and queries a Teacher for paired video and action targets. The Student learns from both targets while retaining its one-step video and action generation at deployment. Across 12 RoboTwin 2.0 tasks, WAM-OPD improves average success from 33.8\% to 65.7\%; across four real-robot tasks, it improves average success from 51.4\% to 64.6\%. With the collected Student histories held fixed, joint video--action supervision achieves the highest observed success on all three ablation tasks, while either modality alone also improves performance. A separate comparison with Teacher-generated histories finds task-dependent differences between the two history sources. These results demonstrate the value of paired video--action supervision for improving WAM policies without increasing their deployed sampling budget.
\end{abstract}
\begingroup
  \noindent\begin{minipage}{\linewidth}
  \setlength{\parskip}{0pt}
  \centering
  \includegraphics[width=0.83\linewidth]{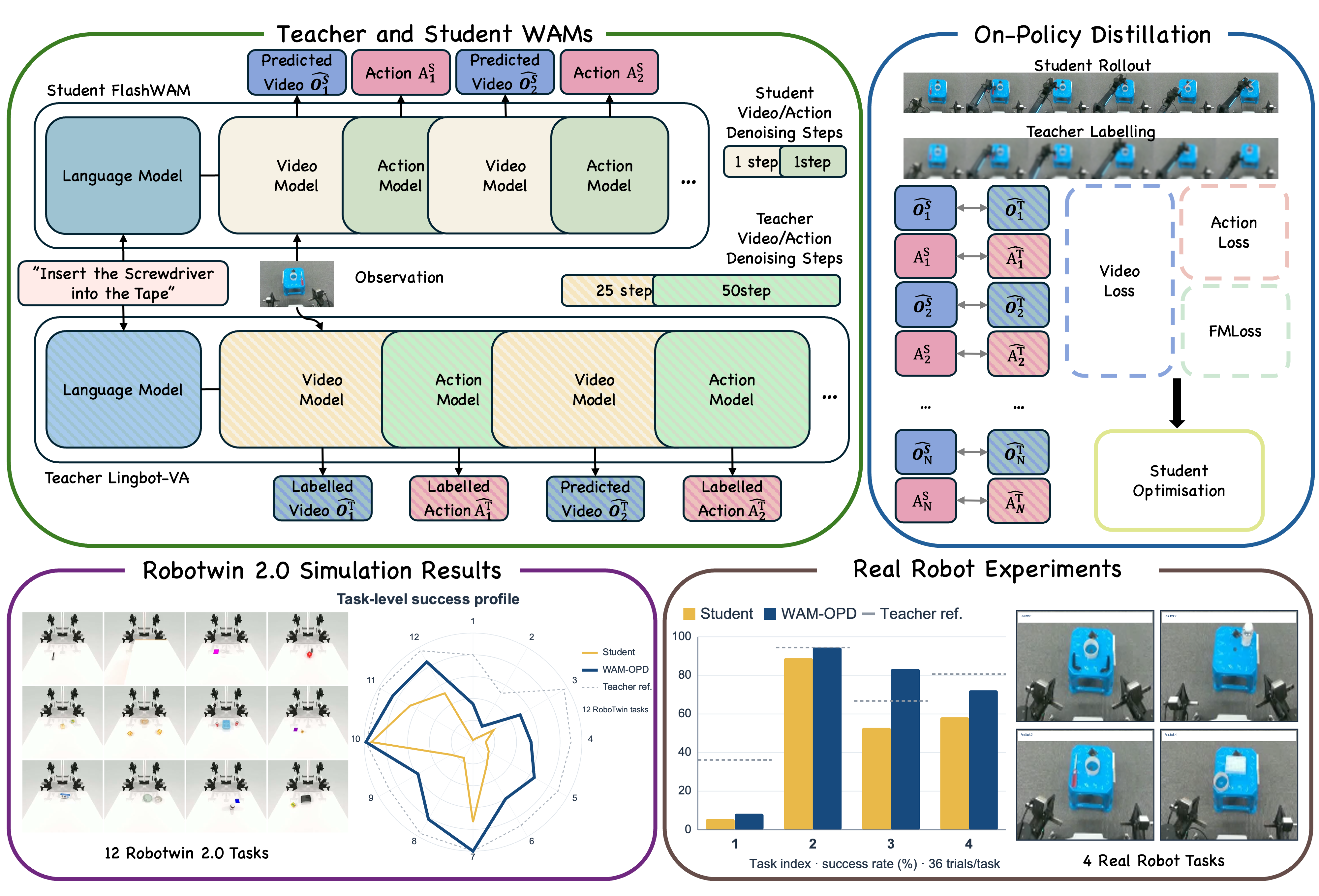}
  \captionof{figure}{Overview of WAM-OPD. On-policy distillation queries the
  Teacher on Student histories and uses paired video and action
  targets to update the Student. The lower panels summarize the evaluation
  settings; task indices follow the row order in Tables~\ref{tab:main-sim-results}
  and~\ref{tab:real-results}.}
  \label{fig:hero}
  \end{minipage}
  \par
\endgroup
\section{Introduction}
\label{sec:introduction}

Robot foundation policies increasingly combine multimodal pretraining with
robot demonstrations, enabling a single policy to control diverse tasks and
embodiments \citep{rt1,rt2,openxembodiment,openvla}. Adapting such a policy to
a new setting involves learning from task-specific experience while retaining
its deployment computation. World Action Models (WAMs) jointly model future
video and robot actions, providing two outputs that can both be supervised
during post-training.

The available training signal, however, depends on how the training contexts
are obtained. Expert teleoperation demonstrations provide paired video and action targets
on the histories they record. Once the adapted policy acts in the
environment, its own decisions can produce histories that differ from those
demonstrations \citep{dagger}. Interactive methods address this distribution
shift by collecting feedback on policy-induced states. Reinforcement learning
can improve behavior from task rewards \citep{simplevlarl2025,pirl2025},
but a task reward does not by itself specify what the WAM should predict for
future video. For WAM post-training, this raises the question of how to
provide paired video and action targets on the histories a policy encounters
during interaction.

We address this question with WAM-OPD (Fig.~\ref{fig:hero}), which applies on-policy distillation to a WAM.
The Student first produces trajectories through environment interaction. A
Teacher then labels the resulting Student histories with video and action
targets, and the Student is updated on those paired targets. This procedure
connects the training histories to the policy being adapted. The Teacher
constructs targets during post-training, while the Student retains its
deployment computation.

Our contribution is a WAM-specific formulation of on-policy distillation:
WAM-OPD constructs paired video and action targets from the Teacher on the same
Student-induced histories, while retaining the Student's video--action
computation used at deployment. We evaluate the method on 12 RoboTwin 2.0
simulation tasks and 4 real-robot tasks, comparing the Student before and
after post-training. Controlled video-only, action-only, and joint-supervision
ablations keep Student histories and the evaluation protocol fixed, while a
separate comparison varies the source of the histories; joint supervision
achieves the highest observed success on the three evaluated ablation tasks.

\section{Related Work}
\label{sec:related}

\paragraph{World Action Models and multimodal robot policies.}
Vision-language-action (VLA) policies map visual observations and language
instructions to robot actions, building on multimodal pretraining and robot
data for general-purpose manipulation \citep{rt1,rt2,openxembodiment,openvla}.
Task-specific manipulation pipelines also use visual detection or sign-language
interfaces~\citep{zhao2024agriculture,signvla2026}.
Diffusion and flow matching provide alternative ways to parameterize continuous
action generation \citep{diffusionpolicy,pi0,flowmatching2023}. World Action
Models (WAMs) extend this setting by modeling future observations together
with action generation. Some methods use predicted visual representations to inform action
generation \citep{videopredictionpolicy}, while others couple video and action
generation within a shared model
\citep{predictionwithaction,unifiedvideoaction,unifiedworldmodels};
autoregressive models such as LingBot-VA support closed-loop control, while
world-model methods such as RoboDreamer use predicted dynamics for imagination
or control \citep{lingbotva2026,robodreamer}. Together,
these approaches use video prediction to support robot action generation.
Our work studies post-training with paired video and action targets on the
histories a policy encounters during execution.

\paragraph{Control and post-training supervision.}
Broader control research studies disturbance rejection \citep{tang2016dobc,tang2019ilc,tang2024slot,tang2026coating}
and multi-agent coordination~\citep{luo2020affine,onuoha2024stress,gong2025distributed}.
RL-based controllers learn policies and value functions for tracking
and stabilization under uncertainty~\citep{tang2024tracking,tang2024stabilization,tang2024delay,bai2026event}.
For pretrained models, post-training methods differ in their learning
signals~\citep{yang2026posttraining} and in the histories used for training. Demonstration-
based supervised fine-tuning uses a fixed collection of expert trajectories.
For WAMs that predict both modalities, these demonstrations provide paired
video and action targets \citep{unifiedworldmodels,lingbotva2026}. Interactive imitation instead queries
an expert on states induced by the learner, as in DAgger \citep{dagger}. Online
reinforcement learning derives updates from rewards or value estimates during
environment interaction; SimpleVLA-RL and $\pi_{\mathrm{RL}}$ study this form of
adaptation for VLA policies \citep{simplevlarl2025,pirl2025}, while WAM-RL
combines action reinforcement learning with reconstruction rewards and online
video SFT \citep{wamrl2026}. Other approaches use imagined or world-model-
generated transitions, predictive critics, or human corrections
\citep{worldsample2026,wcm2026,hiwm2026,wise2026}. These methods broaden the sources of post-training data.
Learned critics can also guide action generation at inference time without
updating the policy~\citep{yang2026guidedactionflow}. WAM-OPD instead
uses a Teacher to provide paired video--action targets on the histories
encountered by the Student.

\paragraph{On-policy distillation.}
On-policy distillation (OPD) supplies supervision on contexts generated by the
policy being trained. In autoregressive language models, the context is a
Student-generated prefix and the Teacher provides a target distribution
\citep{gkd}. Extensions to continuous generators match Teacher transitions or
vector fields at Student-visited sampling states
\citep{diffusionopd2026,flowopd2026}; Vision-OPD applies the principle to
visual reasoning in multimodal language models \citep{visionopd2026}. In
robotics, VLA-OPD transfers Teacher action supervision to environment states
visited by a VLA policy \citep{vlaopd2026}. VLA-OPD provides token-level action
supervision, whereas WAM-OPD operates on a flow-matching WAM whose action
generation is conditioned on its predicted video. These methods share
Student-induced contexts, while WAM-OPD addresses the paired-target setting
of WAMs: on each Student history, the Teacher supplies both video and action
targets, and the Student is trained with the same video--action computation
used at deployment.

\section{Method}
\label{sec:method}

\begin{figure}[htbp]
  \centering
  \includegraphics[width=\linewidth]{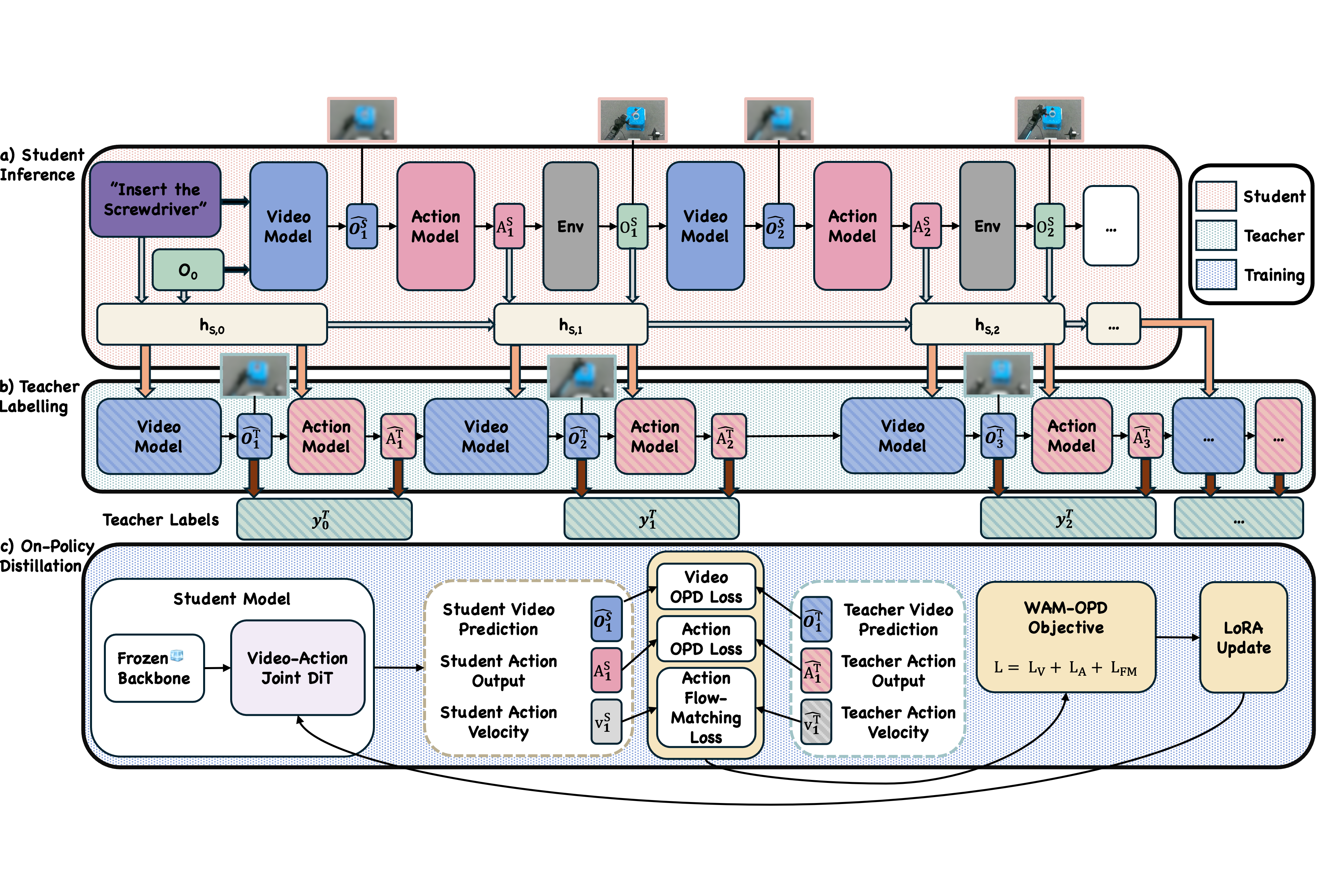}
  \caption{\textbf{WAM-OPD training procedure.} The Student interacts with the
  environment and collects histories $h_{S,i}$. The Teacher is queried on
  these histories to produce paired video and action targets $(\hat O_i^T,
  \hat A_i^T)$. The resulting labeled histories form a fixed training set on
  which the Student recomputes its outputs with the deployed video-to-action
  computation and optimizes the selected losses.}
  \label{fig:method}
\end{figure}

WAM-OPD separates the source of a training context from the source of its
supervision (Fig.~\ref{fig:method}). During rollout, the Student reads its current history, predicts a
video, uses that prediction to generate an action chunk, and executes the
chunk in the environment. The resulting observation extends the history for
the next decision. A Teacher is then queried on these Student-generated
histories to provide video and action targets. The Student is updated on those
targets using the same video-to-action computation used during rollout. We use
$O$ for an environment observation, $\hat O$ for a predicted video, and $A$
for an action chunk.

\subsection{Student histories and Teacher labels}
\label{sec:method-setting}

At decision index $i$, the Student has received a sequence of observations and
has executed a sequence of action chunks. Together with the task instruction,
these quantities define the context on which the policy makes its next
prediction:
\begin{equation}
  h_{S,i}=(O^S_{\leq i},A^S_{<i},\ell),
  \label{eq:history}
\end{equation}
where $O^S_{\leq i}$ and $A^S_{<i}$ denote the observations received and
actions executed during the Student rollout, and $\ell$ is the task
instruction. The Student applies its deployed computation to this context. It
first predicts a video and then conditions action generation on that
prediction:
\begin{align}
  \hat O_i^S &= f^{\mathrm{video}}_\theta(h_{S,i}),
  &
  A_i^S &= f^{\mathrm{action}}_\theta
    \bigl(h_{S,i},\operatorname{sg}(\hat O_i^S)\bigr).
  \label{eq:student-forward}
\end{align}
Here $\operatorname{sg}(\cdot)$ denotes stop-gradient. It ensures that the
video prediction supplied to the action branch is used as an input during the
update without backpropagating through the generated video tensor into its
video-generation graph. Shared adapters used by subsequent conditioning and
action computation remain trainable. After the
Student executes $A_i^S$, the environment returns the next observation; this
observation extends the history used at the next decision index.

The Teacher does not interact with the environment during this stage. For
each history visited by the Student, it is queried on that same context, and
its video and action branches follow the corresponding Teacher computation:
\begin{align}
  \hat O_i^T &= f_T^{\mathrm{video}}(h_{S,i}),
  &
  \hat A_i^T &= f_T^{\mathrm{action}}
    \bigl(h_{S,i},\hat O_i^T\bigr).
  \label{eq:teacher-label}
\end{align}
The Teacher action target is therefore conditioned on the Teacher's own
predicted video, while the Student action prediction is conditioned on the
Student's predicted video in \eqref{eq:student-forward}. This keeps the
Student loss aligned with the computation it will use when deployed. The pair
$y_i^T=(\hat O_i^T,\hat A_i^T)$ is then attached to the Student history and
serves as its supervision. Because the predicted video conditions the action
branch, the video target constrains a signal consumed by the deployed action
computation.

\subsection{Video--action supervision}
\label{sec:modality-supervision}

For every labeled history, the Student recomputes a video prediction, an
action chunk, and an action-generation velocity. We compare the Student's
video prediction with $\hat O_i^T$ and its action chunk with $\hat A_i^T$ using
the elementwise pseudo-Huber discrepancy
$\rho_c(x,y)=\sqrt{(x-y)^2+c^2}-c$. We write $d_{\mathrm{PH}}$ for the mean
of this discrepancy over valid elements. The video term is evaluated in video
token space, and the action term is evaluated in action space.

The action generator is also trained at the level of its flow-matching
velocity. For the interpolation
$x_{A,i}(\sigma)=(1-\sigma)\hat A_i^T+\sigma\epsilon_{A,i}$,
$\sigma\in[0,1]$ moves from the clean Teacher action target to Gaussian noise
$\epsilon_{A,i}$. The corresponding target velocity is
$\hat v_i^T=\epsilon_{A,i}-\hat A_i^T$. We evaluate the Student velocity
$v_i^S$ at the initial action sampling state ($\sigma=1$, native scheduler
timestep $t=1000$), using the same stored action noise, valid elements, and
token positions as in Teacher labeling. Thus,
the action branch receives one term that matches its predicted action and a
second term that matches the velocity used to generate that action.

The loss for supervision mode $\mathcal M$ is
\begin{equation}
\begin{aligned}
\mathcal L_{\mathcal M} ={}& m_v\lambda_v
 d_{\mathrm{PH}}(\hat O_i^S,\hat O_i^T) \\
&+m_a\left[\lambda_a d_{\mathrm{PH}}(A_i^S,\hat A_i^T)
 +\lambda_{\mathrm{FM}}\left\|v_i^S-\hat v_i^T\right\|_2^2\right],
\end{aligned}
\label{eq:objective}
\end{equation}
where $(m_v,m_a)=(0,1)$ for action-only supervision, $(1,0)$ for video-only
supervision, and $(1,1)$ for joint supervision. The switch $m_v$ activates the
video term. The switch $m_a$ activates both action terms, so action-only supervision still trains the action
prediction and its flow-matching velocity. Joint supervision activates all
three terms. All modes retain the complete video-to-action computation in
\eqref{eq:student-forward}; they differ only in which losses contribute to
the update. The velocity discrepancy is averaged over valid action elements.

\subsection{Optimization and training procedure}
\label{sec:optimization}

The procedure has two stages. First, we collect Student rollouts and query the
Teacher on every resulting history, storing the contexts and labels in a fixed
dataset. We partition this dataset into $\mathcal D_{\mathrm{train}}$ for
optimization and $\mathcal D_{\mathrm{cal}}$ for calibration. Second, we
optimize the Student adapters on $\mathcal D_{\mathrm{train}}$; the environment
is not revisited during this optimization stage. Algorithm~\ref{alg:wam-opd}
summarizes these stages.

The base Student and Teacher weights remain frozen, and only rank-8 JointLoRA
adapters in Transformer blocks 0--29 are optimized with AdamW. The reported
runs use three epochs, learning rate $2\times10^{-5}$, effective batch
size 4, gradient-norm clipping at 2, and
$(\lambda_v,\lambda_a,\lambda_{\mathrm{FM}})=(1,1,0.2)$ with pseudo-Huber
scale $c=10^{-3}$. The same optimization procedure is used for the ablated
supervision modes after applying their corresponding loss switches. We reduce
the loss by first averaging losses over labeled decision contexts within each
trajectory and then averaging trajectories equally. The 12 calibration
trajectories are held out from optimization; two decision contexts, one early
and one late, are selected deterministically from each as fixed calibration
anchors. Before training and after every epoch, we
evaluate the video, action, flow-matching, and weighted total losses on these
anchors without gradients to monitor held-out loss.

\begin{algorithm}[htbp]
\caption{WAM-OPD}
\label{alg:wam-opd}
\begin{algorithmic}[1]
\Require Student $\pi_\theta$, frozen Teacher $\pi_T$, environment $E$,
rollout budget $K=36$, supervision mode $\mathcal M$
\Ensure WAM-OPD Student $\pi_{\theta'}$
\State $\mathcal D\gets\emptyset$
\For{$k=1,\ldots,K$}
  \State Roll out $\pi_\theta$ in $E$ and collect Student histories
  $\{h_{S,i}\}$
  \For{each collected history $h_{S,i}$}
    \State Query $\pi_T$ to obtain $(\hat O_i^T,\hat A_i^T)$ and append
    $(h_{S,i},\hat O_i^T,\hat A_i^T)$ to $\mathcal D$
  \EndFor
\EndFor
\State Partition collected trajectories into 24 train trajectories
$\mathcal D_{\mathrm{train}}$ and 12 calibration trajectories
$\mathcal D_{\mathrm{cal}}$
\State Select two fixed early/late anchors per calibration trajectory
\State Evaluate calibration losses on $\mathcal D_{\mathrm{cal}}$ without gradients
\For{each training epoch}
  \For{each minibatch in $\mathcal D_{\mathrm{train}}$}
    \State Compute Student video, detached-video-conditioned action, and action velocity
    \State Evaluate $\mathcal L_{\mathcal M}$ in
    \eqref{eq:objective} and update only the Student adapters
  \EndFor
  \State Re-evaluate calibration losses on $\mathcal D_{\mathrm{cal}}$ without gradients
\EndFor
\State \Return $\pi_{\theta'}$
\end{algorithmic}
\end{algorithm}

\section{Experimental Evaluation}
\label{sec:experiments}

We first evaluate the change in closed-loop task success after applying \method{}
to the Student. We then examine how the two supervision signals contribute to
this change and how the source of training histories affects the update.

\subsection{Experimental Setup}
\label{sec:experimental-setup}

\paragraph{Models and preparation.}
We instantiate \method{} with Flash-WAM as the Student and LingBot-VA as the
Teacher \citep{flashwam2026,lingbotva2026}. The main comparison is between
Flash-WAM and the same Student after WAM-OPD post-training; LingBot-VA is
evaluated as a Teacher reference. For the real-robot experiments, we first
adapt LingBot-VA to the target platform using 40 teleoperated demonstration
trajectories per task collected on the physical robot, obtain the Flash-WAM
Student through distillation, and then apply \method{} to that Student.
The sampling schedules used for all policies are listed in
\Cref{tab:model-configurations}; Flash-WAM and \method{} each use one video
step and one action step in both evaluation settings.

\begin{table}[htbp]
  \centering
  \small
  \caption{Video and action sampling steps used by each policy}
  \label{tab:model-configurations}
  \begin{tabular}{lcc}
    \toprule
    Policy & Video steps & Action steps \\
    \midrule
    LingBot-VA (Teacher reference) & 25 & 50 \\
    Flash-WAM (Student)  & 1  & 1 \\
    \method{} (Ours)     & 1  & 1 \\
    \bottomrule
  \end{tabular}
\end{table}

\paragraph{Tasks and success criteria.}
The simulation study covers 12 RoboTwin 2.0 \texttt{demo\_clean} tasks
\citep{robotwin2}. The real-robot study covers 4 manipulation tasks:
Put Gripper Parts, Shake Bottle, Insert Screw Driver Tape, and Put Tape
Platform. Examples of the evaluation tasks appear in Fig.~\ref{fig:hero}.
The real-robot success criteria are task-specific:
Put Gripper Parts succeeds when both gripper parts are placed inside a tape
roll. Put Tape Platform succeeds when the tape roll is moved from the stool
onto the white platform. Insert Screw Driver Tape succeeds when the
screwdriver is placed inside the tape roll. Shake Bottle succeeds when the
robot lifts the bottle, gently shakes it three times, and returns it to the
stool.

The physical setup uses two wrist-mounted cameras and one head-mounted camera.
Put Gripper Parts is bimanual; Shake Bottle uses the right arm, while Insert
Screw Driver Tape and Put Tape Platform use the left arm. Each task starts from
the same nominal reset configuration, with object placements varying within
approximately 3~cm of their nominal locations. Teleoperation and evaluation
trajectories follow the same reset convention. During inference, action
coordinates for the inactive arm are masked: the left arm for Shake Bottle
and the right arm for Insert Screw Driver Tape and Put Tape Platform.

\begin{figure}[htbp]
  \centering
  \begin{subfigure}{0.98\linewidth}
    \includegraphics[width=\linewidth]{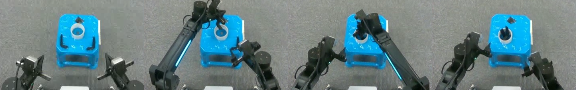}
    \caption{Put Gripper Parts}
  \end{subfigure}\par\vspace{1pt}
  \begin{subfigure}{0.98\linewidth}
    \includegraphics[width=\linewidth]{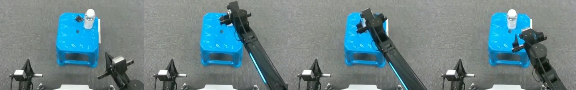}
    \caption{Shake Bottle}
  \end{subfigure}\par\vspace{1pt}
  \begin{subfigure}{0.98\linewidth}
    \includegraphics[width=\linewidth]{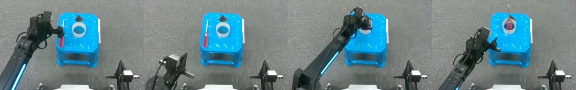}
    \caption{Insert Screw Driver Tape}
  \end{subfigure}\par\vspace{1pt}
  \begin{subfigure}{0.98\linewidth}
    \includegraphics[width=\linewidth]{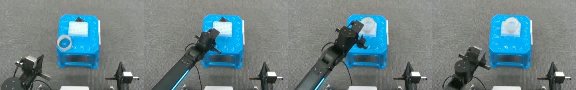}
    \caption{Put Tape Platform}
  \end{subfigure}
  \caption{Representative successful real-robot episodes. Each row shows four
  chronological frames from one task, from the initial configuration to the
  completed episode.}
  \label{fig:real-episodes}
\end{figure}

Fig.~\ref{fig:real-episodes} shows representative successful episodes for
these four tasks. Each strip samples four chronological frames from one
episode; the strips are intended to make the task definitions concrete rather
than to serve as an additional quantitative comparison.

\paragraph{Training and evaluation protocol.}
Unless otherwise stated, \method{} uses the joint video--action objective
and optimization protocol in Section~\ref{sec:method}. The main simulation
setting collects 36 Student rollout trajectories per task: 24 are used for
optimization and 12 are reserved for calibration. Evaluation uses 60
held-out trials per policy and task. Student and WAM-OPD
evaluations use matched task seeds and the same noise bank for paired
comparisons; the simulation ablations use the same paired protocol. Each real-robot
policy is evaluated over 36 rollouts per task, using the same task definitions
and success criteria. The real-robot post-training uses the same 24-trajectory
optimization and 12-trajectory calibration split and the same optimization
budget as the simulation setting.

\paragraph{Metrics.}
The primary metric is task success rate, reported in percent. Averages
weight each task equally. We report the change from Flash-WAM to \method{}
as $\Delta$ in percentage points (PP). All averages and changes are calculated
from success counts before rounding to one decimal place.

\subsection{Main Results}
\label{sec:main-results}

\paragraph{Simulation.}
\Cref{tab:main-sim-results} compares the Teacher reference, Flash-WAM, and
\method{} across the 12-task simulation suite. The collection budget is 36
Student trajectories per task, split into 24 optimization and 12 calibration
trajectories.

\begin{table}[htbp]
  \centering
  \small
  \caption{Task success rates (\%) on 12 RoboTwin 2.0 tasks, evaluated over
  60 trials per policy and task}
  \label{tab:main-sim-results}
  \setlength{\tabcolsep}{3pt}
  \begin{tabular*}{\textwidth}{@{\extracolsep{\fill}}lcccc}
    \toprule
    Task & \textcolor{teachergray}{\shortstack{Teacher\\(Reference)}}
      & \cellcolor{studentgold}\shortstack{Flash-WAM\\(Student)}
      & \cellcolor{ourbluehead}\shortstack{\textbf{\method{}}\\\textbf{(Ours)}}
      & \cellcolor{ourbluehead}\textcolor{gainblue}{$\Delta$ (PP)} \\
    \midrule
    \texttt{handover\_mic} & \textcolor{teachergray}{80.0} & \cellcolor{studentgold!35}1.7 & \cellcolor{ourblue!55}\textcolor{gainblue}{35.0} & \cellcolor{ourblue!35}\textcolor{gainblue}{\textbf{+33.3}} \\
    \texttt{put\_object\_cabinet} & \textcolor{teachergray}{51.7} & \cellcolor{studentgold!35}3.3 & \cellcolor{ourblue!55}\textcolor{gainblue}{16.7} & \cellcolor{ourblue!35}\textcolor{gainblue}{\textbf{+13.3}} \\
    \texttt{place\_fan} & \textcolor{teachergray}{96.7} & \cellcolor{studentgold!35}21.7 & \cellcolor{ourblue!55}\textcolor{gainblue}{50.0} & \cellcolor{ourblue!35}\textcolor{gainblue}{\textbf{+28.3}} \\
    \texttt{place\_a2b\_left} & \textcolor{teachergray}{90.0} & \cellcolor{studentgold!35}11.7 & \cellcolor{ourblue!55}\textcolor{gainblue}{53.3} & \cellcolor{ourblue!35}\textcolor{gainblue}{\textbf{+41.7}} \\
    \texttt{place\_a2b\_right} & \textcolor{teachergray}{90.0} & \cellcolor{studentgold!35}16.7 & \cellcolor{ourblue!55}\textcolor{gainblue}{65.0} & \cellcolor{ourblue!35}\textcolor{gainblue}{\textbf{+48.3}} \\
    \texttt{place\_bread\_basket} & \textcolor{teachergray}{88.3} & \cellcolor{studentgold!35}25.0 & \cellcolor{ourblue!55}\textcolor{gainblue}{60.0} & \cellcolor{ourblue!35}\textcolor{gainblue}{\textbf{+35.0}} \\
    \texttt{place\_cans\_plasticbox} & \textcolor{teachergray}{100.0} & \cellcolor{studentgold!35}73.3 & \cellcolor{ourblue!55}\textcolor{gainblue}{100.0} & \cellcolor{ourblue!35}\textcolor{gainblue}{\textbf{+26.7}} \\
    \texttt{stamp\_seal} & \textcolor{teachergray}{96.7} & \cellcolor{studentgold!35}16.7 & \cellcolor{ourblue!55}\textcolor{gainblue}{81.7} & \cellcolor{ourblue!35}\textcolor{gainblue}{\textbf{+65.0}} \\
    \texttt{rotate\_qrcode} & \textcolor{teachergray}{88.3} & \cellcolor{studentgold!35}23.3 & \cellcolor{ourblue!55}\textcolor{gainblue}{58.3} & \cellcolor{ourblue!35}\textcolor{gainblue}{\textbf{+35.0}} \\
    \texttt{place\_container\_plate} & \textcolor{teachergray}{98.3} & \cellcolor{studentgold!35}93.3 & \cellcolor{ourblue!55}\textcolor{gainblue}{98.3} & \cellcolor{ourblue!35}\textcolor{gainblue}{\textbf{+5.0}} \\
    \texttt{move\_pillbottle\_pad} & \textcolor{teachergray}{98.3} & \cellcolor{studentgold!35}66.7 & \cellcolor{ourblue!55}\textcolor{gainblue}{85.0} & \cellcolor{ourblue!35}\textcolor{gainblue}{\textbf{+18.3}} \\
    \texttt{place\_object\_stand} & \textcolor{teachergray}{96.7} & \cellcolor{studentgold!35}51.7 & \cellcolor{ourblue!55}\textcolor{gainblue}{85.0} & \cellcolor{ourblue!35}\textcolor{gainblue}{\textbf{+33.3}} \\
    \midrule
    \textbf{Average} & \textcolor{teachergray}{89.6} & \cellcolor{studentgold!35}\textbf{33.8} & \cellcolor{ourblue!55}\textcolor{gainblue}{\textbf{65.7}} & \cellcolor{ourblue!35}\textcolor{gainblue}{\textbf{+31.9}} \\
    \bottomrule
  \end{tabular*}
\end{table}

\method{} increases average success from 33.8\% to 65.7\%, a gain of
31.9 percentage points, with higher success on all 12 tasks. The largest gain
is on Stamp Seal, from 16.7\% to 81.7\%. Improvements also extend to tasks
with high initial success: Place Container Plate rises from 93.3\% to
98.3\%. The gains therefore occur across tasks with different initial Student
performance, while the post-trained policy retains one-step video and action
generation.
Across the evaluated simulation tasks, larger Teacher--Student success gaps
often coincide with larger WAM-OPD gains.

\paragraph{Real robot.}
We next test the same post-training procedure on the physical robot.
\Cref{tab:real-results} reports success on the four manipulation tasks defined
above.

\begin{table}[htbp]
  \centering
  \small
  \caption{Real-robot task success rates (\%) over 36 evaluation rollouts per
  policy and task}
  \label{tab:real-results}
  \setlength{\tabcolsep}{4pt}
  \begin{tabular}{lcccc}
    \toprule
    Task & \textcolor{teachergray}{\shortstack{Teacher\\(Reference)}}
      & \cellcolor{studentgold}\shortstack{Flash-WAM\\(Student)}
      & \cellcolor{ourbluehead}\shortstack{\textbf{\method{}}\\\textbf{(Ours)}}
      & \cellcolor{ourbluehead}\textcolor{gainblue}{$\Delta$ (PP)} \\
    \midrule
    \shortstack{Put Gripper\\Parts} & \textcolor{teachergray}{36.1} & \cellcolor{studentgold!35}5.6 & \cellcolor{ourblue!55}\textcolor{gainblue}{8.3} & \cellcolor{ourblue!35}\textcolor{gainblue}{\textbf{+2.8}} \\
    Shake Bottle & \textcolor{teachergray}{94.4} & \cellcolor{studentgold!35}88.9 & \cellcolor{ourblue!55}\textcolor{gainblue}{94.4} & \cellcolor{ourblue!35}\textcolor{gainblue}{\textbf{+5.6}} \\
    \shortstack{Insert Screw\\Driver Tape} & \textcolor{teachergray}{66.7} & \cellcolor{studentgold!35}52.8 & \cellcolor{ourblue!55}\textcolor{gainblue}{83.3} & \cellcolor{ourblue!35}\textcolor{gainblue}{\textbf{+30.6}} \\
    \shortstack{Put Tape\\Platform} & \textcolor{teachergray}{80.6} & \cellcolor{studentgold!35}58.3 & \cellcolor{ourblue!55}\textcolor{gainblue}{72.2} & \cellcolor{ourblue!35}\textcolor{gainblue}{\textbf{+13.9}} \\
    \midrule
    \textbf{Average} & \textcolor{teachergray}{69.4} & \cellcolor{studentgold!35}\textbf{51.4} & \cellcolor{ourblue!55}\textcolor{gainblue}{\textbf{64.6}} & \cellcolor{ourblue!35}\textcolor{gainblue}{\textbf{+13.2}} \\
    \bottomrule
  \end{tabular}
\end{table}

The average success rate increases from 51.4\% to 64.6\%, a gain of
13.2 percentage points, with higher observed success on all four tasks.
The largest gain is on Insert Screw Driver Tape, from 52.8\% to 83.3\%.
Put Gripper Parts remains the most challenging task: complete placement
increases from 5.6\% to 8.3\%. The improvement therefore also holds for physical execution, with its
magnitude varying across tasks.

\subsection{Supervision Composition}
\label{sec:supervision-ablation}

\begin{figure}[htbp]
  \centering
  \includegraphics[width=0.64\linewidth]{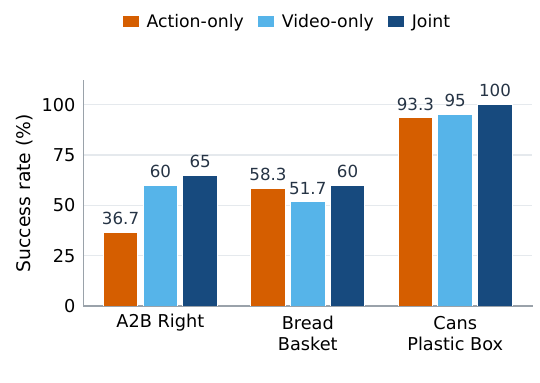}
  \caption{Effect of supervision composition on task success (\%). Each
  setting uses the same 36-trajectory collection split (24 optimization, 12
  calibration) and 60 matched evaluation trials. Only the active loss terms differ. Among these modes, joint
  supervision yields the highest observed success rate on all three tasks.}
  \label{fig:supervision-ablation}
\end{figure}

We compare action-only, video-only, and joint supervision to examine the
contribution of each Teacher output. Student histories, Teacher labels,
initialization, adapter capacity, optimization budget, and evaluation are held
fixed. In \eqref{eq:objective}, action-only sets
$(m_v,m_a)=(0,1)$, retaining both action losses; video-only sets
$(m_v,m_a)=(1,0)$, disabling both action losses; and joint supervision sets
$(m_v,m_a)=(1,1)$. \Cref{fig:supervision-ablation} reports the resulting task
success rates.

Joint supervision achieves the highest observed success rate on all three
tasks, exceeding the better single-modality setting by 5.0, 1.7, and 5.0
percentage points on Place A2B Right, Place Bread Basket, and Place Cans
Plastic Box, respectively. The relative ordering of the two single-modality
settings varies: video-only is higher on the first and third tasks, whereas
action-only is higher on Place Bread Basket. These results are consistent with a benefit from combining both supervision
signals across these tasks.

Both single-modality settings improve over Flash-WAM on all three tasks.
Action-only supervision improves success by 20.0, 33.3, and 20.0 percentage
points, while video-only supervision improves it by 43.3, 26.7, and 21.7
percentage points, respectively. Action-only supervision directly targets robot behavior. The video-only
setting also improves task success even when both action losses are disabled. Together
with the joint results, these comparisons are consistent with using both video
and action targets for WAM post-training.

\subsection{Source of Training Histories}
\label{sec:data-efficiency}

We compare two sources of training histories under the same joint supervision
objective: histories collected from Student rollouts and histories collected
from Teacher rollouts. Both sources use 36 trajectories per task, split into
24 optimization and 12 calibration trajectories, using the same task-specific
rollout seeds and prompts. The optimizer, learning rate, effective batch size,
number of training epochs, and evaluation protocol are held fixed.
This matches the rollout and optimization configuration so that the comparison
focuses on the source of the training histories.
\Cref{tab:history-support} compares the two sources on the same three tasks
as the supervision ablation.

\begin{table}[htbp]
  \centering
  \small
  \caption{Effect of training-history source on task success (\%), evaluated
  over 60 matched trials per task}
  \label{tab:history-support}
  \setlength{\tabcolsep}{4pt}
  \begin{tabular}{lccc}
    \toprule
    Task & \cellcolor{studentgold}\shortstack{Flash-WAM\\(Student)}
      & \cellcolor{offlinegray}\shortstack{Joint / Teacher\\histories}
      & \cellcolor{ourbluehead}\shortstack{Joint / Student\\histories} \\
    \midrule
    \texttt{place\_a2b\_right} & \cellcolor{studentgold!35}16.7
      & \cellcolor{offlinegray}26.7
      & \cellcolor{ourblue!55}\textcolor{gainblue}{65.0} \\
    \texttt{place\_bread\_basket} & \cellcolor{studentgold!35}25.0
      & \cellcolor{offlinegray}75.0
      & \cellcolor{ourblue!55}\textcolor{gainblue}{60.0} \\
    \texttt{place\_cans\_plasticbox} & \cellcolor{studentgold!35}73.3
      & \cellcolor{offlinegray}100.0
      & \cellcolor{ourblue!55}\textcolor{gainblue}{100.0} \\
    \bottomrule
  \end{tabular}
\end{table}

Student histories yield higher success on Place A2B Right (65.0\% versus
26.7\%), while Teacher histories yield higher success on Place Bread Basket
(75.0\% versus 60.0\%). Both reach 100.0\% on Place Cans Plastic Box.
Both sources improve over Flash-WAM on all three tasks; their relative
performance is task dependent.

\section{Conclusion and Future Work}
\label{sec:conclusion}

WAM-OPD demonstrates the value of supervising both video prediction and action
generation during WAM post-training. The ablations show that either signal can
improve task performance, but the more effective signal varies by task. With
Student histories held fixed, joint supervision achieves the highest success
rate on all three tasks in the supervision ablation. In simulation and on the
physical robot, WAM-OPD improves task success while retaining the Student's
inference budget.
Both Student and Teacher rollout histories improve over Flash-WAM in the
history-source comparison, with neither source consistently outperforming
the other.

These findings motivate exploring how Teacher targets can complement task
rewards in reinforcement learning. Video and action targets could serve as
auxiliary objectives while rewards guide policy improvement toward task
success. We also plan to extend WAM-OPD to other WAM architectures and smaller
Student policies. The same formulation may extend to flow-matching VLA
policies with coupled visual--action computation. Beyond WAM-to-WAM
supervision, transferring targets between WAM and VLA policies could connect
their different prediction capabilities:
WAM-generated video and actions could support VLA training, while VLA action
predictions could supervise the action component of a WAM.

{\sloppy
\bibliographystyle{IEEEtran}
\bibliography{references}
}
\end{document}